# Transformer Meets Convolution: A Bilateral Awareness Network for Semantic Segmentation of Very Fine Resolution Urban Scene Images

**Libo Wang [1, †], Rui Li [1, †], Dongzhi Wang [2, *], Chenxi Duan [3], Teng Wang [2] and Xiaoliang Meng [1]**

[1] School of Remote Sensing and Information Engineering, Wuhan University, Wuhan 430079, China; wanglibo@whu.edu.cn (L.W.); lironui@whu.edu.cn (R.L.); xmeng@whu.edu.cn (X.M.)

[2] Surveying and Mapping Institute Lands and Resource Department of Guangdong Province, Guangzhou 510500, China; wangteng43@hotmail.com (T.W.)

[3] State Key Laboratory of Information Engineering in Surveying, Mapping, and Remote Sensing, Wuhan University, Wuhan 430079, China; chenxiduan@whu.edu.cn (C.D.)

* Correspondence: harwang1984@foxmail.com (D.W.); Tel: 020-38334381

† Equal contribution.

**Abstract:** Semantic segmentation from very fine resolution (VFR) urban scene images plays a significant role in several application scenarios including autonomous driving, land cover classification, and urban planning, etc. However, the tremendous details contained in the VFR image, especially the considerable variations in scale and appearance of objects, severely limit the potential of the existing deep learning approaches. Addressing such issues represents a promising research field in the remote sensing community, which paves the way for scene-level landscape pattern analysis and decision making. In this paper, we propose a Bilateral Awareness Network which contains a dependency path and a texture path to fully capture the long-range relationships and fine-grained details in VFR images. Specifically, the dependency path is conducted based on the ResT, a novel Transformer backbone with memory-efficient multi-head self-attention, while the texture path is built on the stacked convolution operation. Besides, using the linear attention mechanism, a feature aggregation module is designed to effectively fuse the dependency features and texture features. Extensive experiments conducted on the three large-scale urban scene image segmentation datasets, i.e., ISPRS Vaihingen dataset, ISPRS Potsdam dataset, and UAVid dataset, demonstrate the effectiveness of our BANet. Specifically, a 64.6% mIoU is achieved on the UAVid dataset. Code is available at https://github.com/WangLibo1995/GeoSeg.

## 1. Introduction

Semantic segmentation of very fine resolution (VFR) urban scene images comprises a hot topic in the remote sensing community [1-6]. It plays a crucial role in various urban applications, such as urban planning [7], vehicle monitoring[8], land cover mapping [9], and change detection [10], building and road extraction [11,12], as well as other practical applications [13-15]. The goal of semantic segmentation is to label each pixel with a certain category. Since geo-objects in urban areas are characterized by large within-class and small between-class variance commonly, semantic segmentation of very fine resolution RGB imagery remains a challenging issue [16,17]. For example, urban buildings made of diverse materials show variant spectral signatures, while buildings and roads made of the same material (e.g. cement) exhibit similar textural information in RGB images.

Due to the advantage in local texture extraction, many researchers have investigated the challenging urban scene segmentation task based on deep convolutional neural networks (DCNNs) [18,19]. Especially, the methods based on fully convolutional neural network (FCN) [20], which can be trained end-to-end, have achieved great breakthroughs in urban scene labelling [21]. In comparison with the traditional machine learning methods, such as support vector machine (SVM) [22], random forest [23], and conditional random field (CRF) [24], the FCN-based methods have demonstrated remarkable generalization capability and high efficiency [25,26]. Therefore, numerous specially designed FCN-based networks have been spawned for urban scene segmentation, including UNet and its variants [4,16,27,28], multi-scale context aggregation networks [29,30], and multi-level feature fusion networks [5], attention-based networks [3,31,32], as well as lightweight networks [33]. For example, Sherrah [21] introduced the FCN to semantically label remote sensing images. Kampffmeyer et al. [34] quantified the uncertainty in urban remote sensing images at the pixel level, thereby enhancing the accuracy of relatively small objects (e.g., Cars). Maggiori et al. [35] designed an auxiliary CNN to learn the features fusion schemes. Multi-modal data were further utilized by Audebert et al. [36] in their V-FuseNet to enhance the segmentation performance. However, when if either modality is unavailable in the test phase caused by sensors' corruption or thick cloud cover [37], such a multi-modal data fusion scheme will be invalid. Kampffmeyer et al. [38], therefore, proposed a hallucination network aiming to replace missing modalities during testing. Besides, enhancing the segmentation accuracy by optimizing object boundaries is another burgeoning research area [39,40].

The accuracy of FCN-based networks, although encouraging, appears to be incompetent for VFR segmentation. The reason is that almost all FCN-based networks are built on DCNNs, while the latter is designed for extracting local patterns and lacks the ability to model global context in its nature [41]. Hence, extensive investigations have been devoted to addressing the above issue since the long-range dependency is vital for segmenting confusing manmade objects in urban areas. Typical methods include dilated convolutional networks which are designed for enlarging the receptive field [42,43] and attentional networks that are proposed for capturing long-range relational semantic content of feature maps [31,44]. Nevertheless, these two networks have never been able to get rid of the dependence on the convolution operation, impairing the effectiveness of long-range information extraction.

Most recently, with its strong ability in long-range dependency capture and sequence-based image modelling, an entirely novel architecture named Transformer [45] has become prominent in various computer vision tasks, such as image classification [46], object detection [47], and semantic segmentation [48]. The schematic flowchart of the Transformer is illustrated in Figure 1 (a). First, the Transformer deploys a patch partition to split the 2D input image into non-overlapping image patches. (H, W) and C denotes the resolution and the channel dimension of the input image, respectively. (P, P) is the resolution of each image patch. Then, a flatten operation and a linear projection are employed to produce the 1D sequence. The length of the sequence is N, where N=(H×W)/P². M is the output dimension of the linear projection. Finally, the sequence is fed into stacked transformer blocks to extract features with long-range dependencies. As shown in Figure 1 (b), a standard transformer block is composed of multi-head self-attention (MHSA) [45], layer norm (LN) [49] and multilayer perceptron (MLP) as well as two addition operations. L represents the number of transformer blocks. Benefiting from the non-convolution structure and attention mechanism, Transformer could capture long-range dependencies more effectively [50].

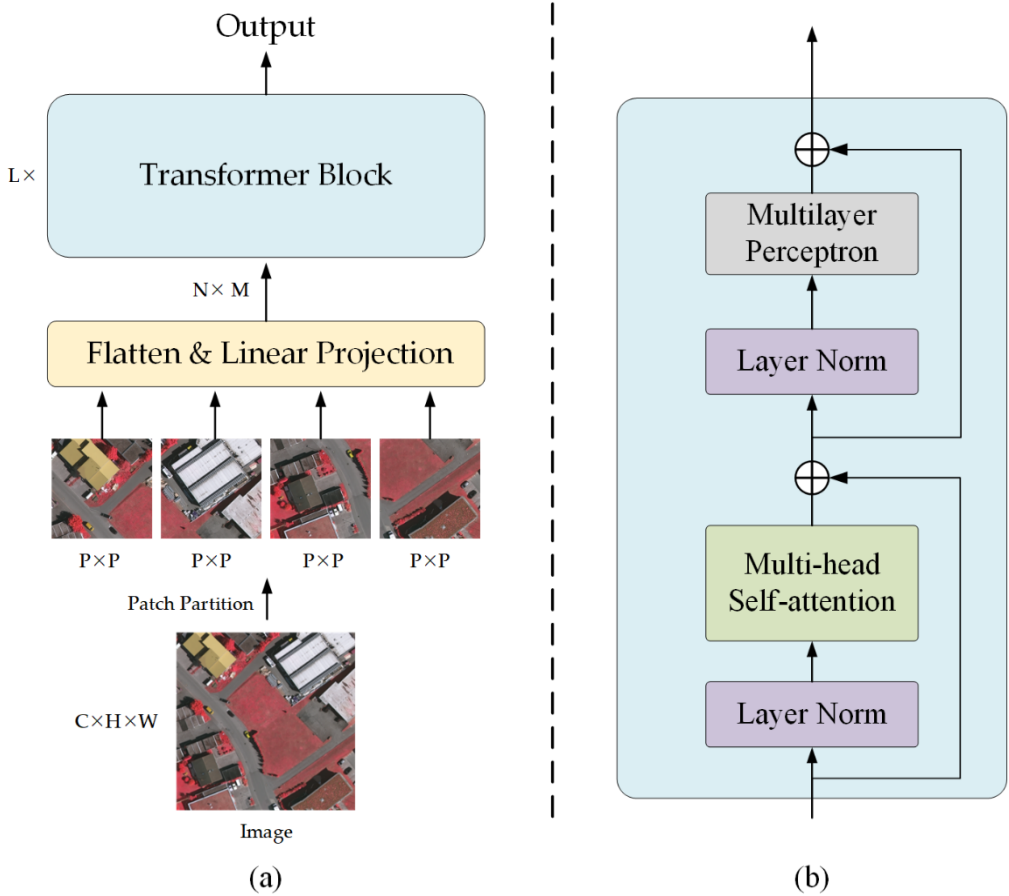

(a)          (b)

**Figure 1.** (a) Illustration of the schematic flowchart of the Transformer. (b) Illustration of a standard transformer block.

Inspired by the advancement of Transformer, in this paper, we propose a Bilateral Awareness Network (BANet) for accurate semantic segmentation of VFR urban scene images. Different from the traditional single-path convolutional neural networks, BANet addresses the challenging urban scene segmentation by constructing two feature extraction paths, as illustrated in Figure 2. Specifically, a texture path using stacked convolution layers is developed to extract the textural feature. Meanwhile, a dependency path using Transformer blocks is established to capture the long-range dependent feature. To leverage the benefits provided by the two features, we design a feature aggregation module (FAM) which introduces the linear attention mechanism to reduce the fitting residual of fused features, thereby strengthening the generalization capability of the network. Experimental results on three large-scale urban scene image segmentation datasets demonstrate the effectiveness of our BANet. Besides, the well-designed bilateral structure could provide a unified solution for semantic segmentation, object detection, and change detection, which undoubtedly boosts deep learning techniques in the remote sensing domain. To sum up, the main contributions of this paper are the following:

(1) A novel bilateral structure composed of convolution layers and transformer blocks is proposed for understanding and labelling very fine resolution urban scene images. It provides a new perspective for capturing textural information and long-range dependencies simultaneously in a single network.

(2) A feature aggregation module is developed to fuse the textural feature and long-range dependent feature extracted by the bilateral structure. It employs linear attention to reduce the fitting residual and greatly improves the generalization of fused features.

The remainder of this paper is organized as follows. The architecture of BANet and its components are detailed in Section 2. Experimental comparisons on three semantic segmentation datasets (UAVid, ISPRS Vaihingen, and Potsdam) are provided in Section 3. A comprehensive discussion is presented in Section 4. Finally, conclusions are drawn in Section 5.

## 2. Bilateral Awareness Network

### 2.1. Overview

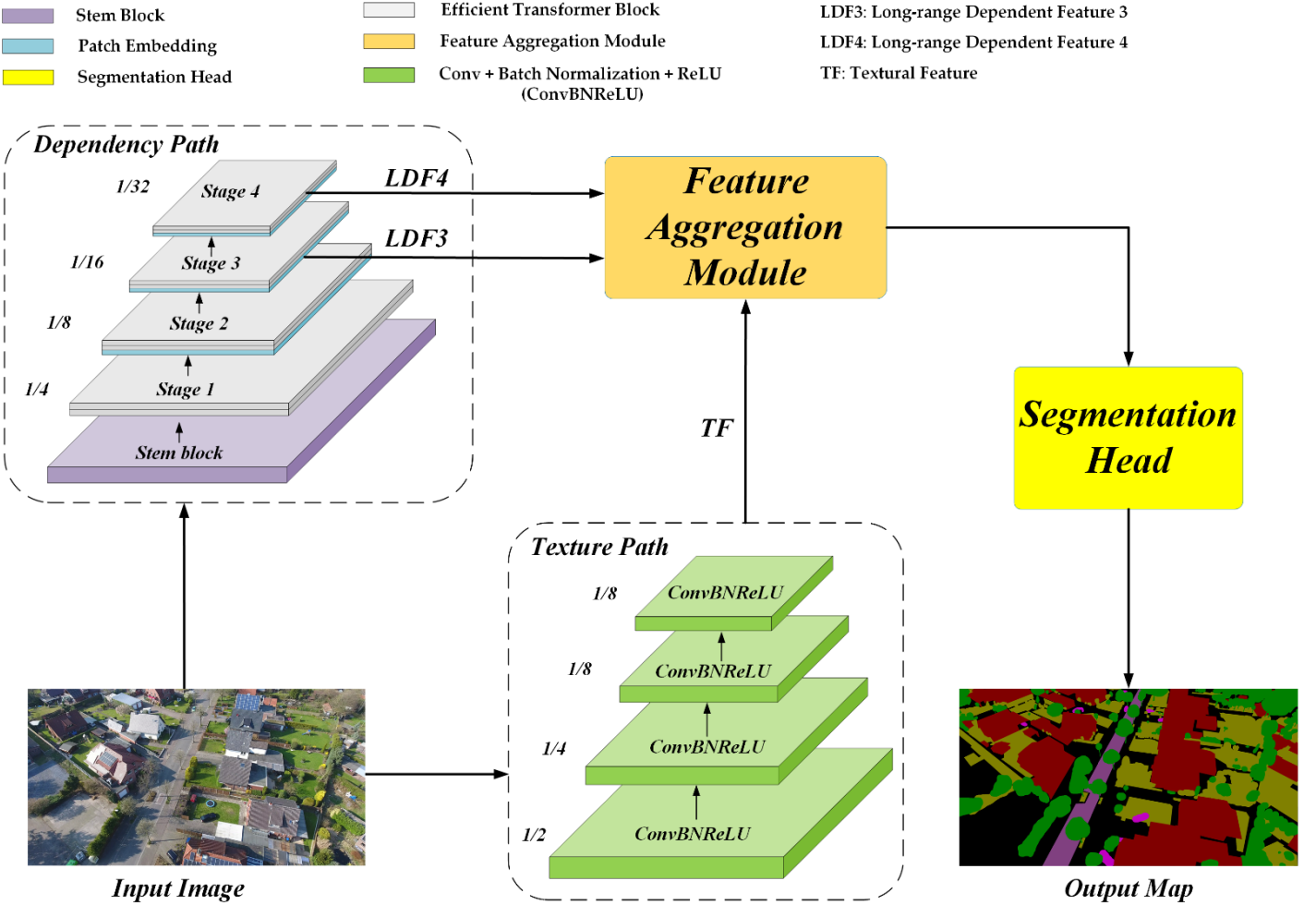

**Figure 2.** The overall architecture of Bilateral Awareness Network (BANet).

The overall architecture of the Bilateral Awareness Network (BANet) is exhibited in Figure 2, where the input image is fed into the dependency path and texture path simultaneously.

The dependency path employs a stem block and four transformer stages (i.e., Stage 1-4) to extract long-range dependent features. Each stage consists of two efficient transformer blocks (ETB). In particular, Stage 2, Stage 3, and Stage 4 involve patch embedding (PE) operations additionally. Proceed by the dependency path, two long-range dependent features (i.e., LDF3 and LDF4) are generated.

The texture path deploys four convolution layers to capture the textural feature (TF), while each convolutional layer is equipped with batch normalization (BN) [51] and ReLU activation function [52]. The downsampling factor is set as 8 for the texture path to preserve spatial details.

Since the outputs of the dependency path and the texture path are in disparate domains, FAM is proposed to merge them effectively. Whereafter, a segmentation head module is attached to convert the fused feature into a segmentation map.

### 2.2. Dependency path

The dependency path is constructed by the ResT-Lite [53] pertained on ImageNet. As an efficient vision transformer, ResT-Lite is suitable for urban scene interpretation due to its balanced trade-off between segmentation accuracy and computational complexity. The main basic modules of the ResT-lite include the stem block, patch embedding and efficient transformer block.

*Stem block:* The stem block aims to shrink the height and width dimension and expand the channel dimension. To capture low-level information effectively, it introduces three 3×3 convolution layers with strides of [2, 1, 2]. The first two convolution layers are followed by BN and ReLU. Proceed by the stem block, the spatial resolution is downscaled by a factor of 4, and the channel dimension is extended from 3 to 64.

*Patch embedding:* The patch embedding aims to down-sampled the feature map for hierarchical feature representation. The output for each patch embedding can be formalized as:

$$PE(\mathbf{X}') = Sigmoid\big(DWConv(\mathbf{X}')\big) \cdot \mathbf{X}' \tag{1}$$

$$\mathbf{X}' = BN(W_s \cdot \mathbf{X}) \tag{2}$$

where $W_s$ represents a convolution layer with a kernel size of s+1 and a stride of s. Here, s is set as 2. DWConv denotes a 3×3 depth-wise convolution [54] with a stride of 1.

*Efficient transformer block:* Each efficient transformer is composed of efficient multi-head self-attention (EMSA) [53], MLP and LN. The output for each efficient transformer block can be formalized as:

$$ETB(\mathbf{X}) = G(\mathbf{X}) + MLP\big(LN\big(G(\mathbf{X})\big)\big) \tag{3}$$

$$G(\mathbf{X}) = \mathbf{X} + EMSA(\mathbf{Q}, \mathbf{K}, \mathbf{V}) \tag{4}$$

The EMSA, a revised self-attention module for computer vision based on MHSA, is the main module of ETB. As illustrated in Figure 3, the detailed steps of EMSA are as follows:

(1) EMSA obtains three vectors $\mathbf{Q}, \mathbf{K}, \mathbf{V}$ from the input vector $\mathbf{X} \in \mathbb{R}^{N \times D}$. Different from the standard multi-head self-attention, EMSA first deploys a depth-wise convolution with a kernel size of r+1 and stride of r to decrease the resolution of $\mathbf{K}$ and $\mathbf{V}$, thereby compressing the computation and memory. For the four transformer stages, r is set as 8, 4, 2, 1, respectively.

(2) To be specific, the input vector $\mathbf{X}$ is reshaped to a new vector with a shape of $D \times H \times W$, where $H \times W = N$. Proceed by the depth-wise convolution, the new vector is reshaped to $D \times h \times w$. Here, $h = H/r$ and $w = W/r$. Then, the new vector is recovered to $n \times D$ as the input of LN, where $h \times w = n$. Thus, the initial shape of $\mathbf{K}$ and $\mathbf{V}$ is $n \times D$. The initial shape of $\mathbf{Q}$ is $N \times D$.

(3) The three vectors $\mathbf{Q}, \mathbf{K}, \mathbf{V}$ are fed into three linear projections and reshaped to $k \times N \times m$, $k \times m \times n$ and $k \times n \times m$, respectively. Here, $k$ denotes the number of heads, m denotes the head dimension, $k \times m = D$.

(4) A matrix multiplication operation is applied on $\mathbf{Q}$ and $\mathbf{K}$ to generate an attention map with the shape of $k \times N \times n$.

(5) The attention map is further proceeded by a convolution layer, a Softmax activation function and an Instance Normalization [55] operation.

(6) A matrix multiplication operation is applied on the proceeded attention map and **V**. Finally, a linear projection is utilized to generate the output vector. The formalization of EMSA can be referred to the equation (5).

$$\text{EMSA}(\mathbf{Q}, \mathbf{K}, \mathbf{V}) = \text{LP}\left(\text{IN}\left(\text{Softmax}\left(\text{Conv}\left(\frac{\mathbf{Q}\mathbf{K}^{\text{T}}}{\sqrt{m}}\right)\right)\right) \cdot \mathbf{V}\right) \tag{5}$$

Here, Conv is a standard 1×1 convolution with a stride of 1. IN denotes an instance normalization operation. LP represents a linear projection that keeps a dimension of $D$.

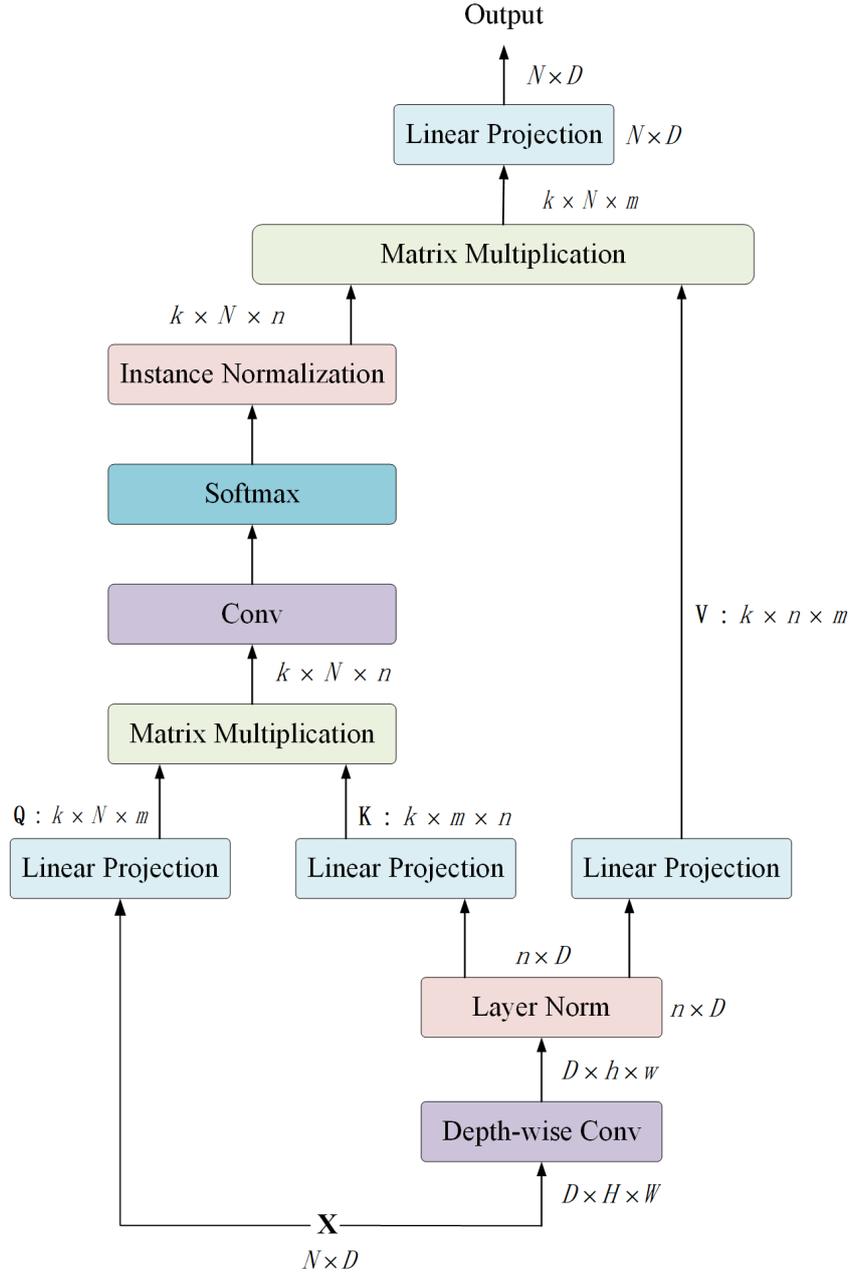

**Figure 3.** The flowchart of efficient multi-head self-attention.

*2.3. Texture path*

The texture path is a lightweight convolutional network, which builds four diverse convolutional layers to capture textural information. The output for the texture path can be formalized as:

$$\mathrm{TF}(\mathbf{X}) = \mathrm{T}_4\left(\mathrm{T}_3\left(\mathrm{T}_2\big(\mathrm{T}_1(\mathbf{X})\big)\right)\right) \tag{6}$$

Here, T represents a combined function consisting of a convolutional layer, a batch normalization operation, and a ReLU activation. The convolutional layer of $\mathrm{T}_1$ has a kernel size of 7 and a stride of 2, which expands the channel dimension from 3 to 64. For $\mathrm{T}_2$ and $\mathrm{T}_3$, the kernel size and stride are 3 and 2, respectively. The channel dimension is kept as 64. For $\mathrm{T}_4$, the convolutional layer is a standard $1 \times 1$ convolution with a stride of 1, expanding the channel dimension from 64 to 128. Thus, the output textural feature is downscaled 8 times and has a channel dimension of 128.

### 2.4. Feature aggregation module

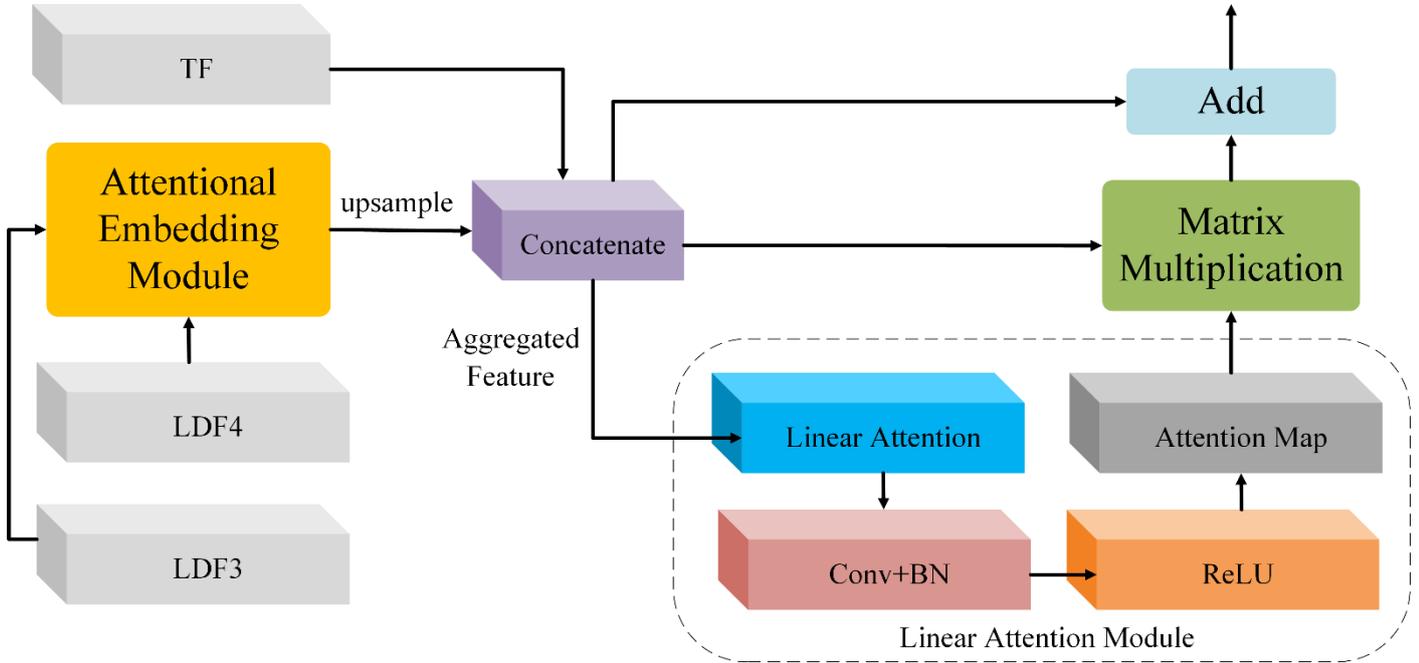

**Figure 4.** The feature aggregation module.

The FAM aims to leverage the benefits of the dependent features and texture features comprehensively for powerful feature representation. As shown in Figure 4, the input features for the FAM include the LDF3, LDF4 and TF. To fuse those features, we first employ an attentional embedding module (AEM) to merge the LDF3 and LDF4. Thereafter, the merged feature is up-sampled to concatenate with the TF, obtaining the aggregated feature. Finally, the linear attention module is deployed to reduce the fitting residual of the aggregated feature (AF). The pipeline of FAM can be denoted as:

$$\mathrm{FAM}(\mathbf{AF}) = \mathbf{AF} \cdot \mathrm{LAM}(\mathbf{AF}) + \mathbf{AF} \tag{7}$$

$$\mathrm{AF}(\mathbf{TF}, \mathbf{LDF3}, \mathbf{LDF4}) = \mathrm{C}\big(\mathrm{U}\big(\mathrm{AEM}(\mathbf{LDF3}, \mathbf{LDF4})\big), \mathbf{TF}\big) \tag{8}$$

Here, C represents the concatenate function. U denotes an upsampling operation with a scale factor of 2. The details of LAM and AEM are as follows.

*Linear attention module*: The conventional dot-product attention mechanism can be defined as:

$$D(\mathbf{Q}, \mathbf{K}, \mathbf{V}) = \rho(\mathbf{Q}\mathbf{K}^{\mathrm{T}})\mathbf{V}. \tag{9}$$

$$\rho(\mathbf{Q}\mathbf{K}^{\mathrm{T}}) = softmax_{row}(\mathbf{Q}\mathbf{K}^{\mathrm{T}}), \tag{10}$$

where query matrix $\mathbf{Q}$, the key matrix $\mathbf{K}$, and the value matrix $\mathbf{V}$ are generated by the corresponding standard $1{\times}1$ convolutional layer with a stride of 1 and $softmax_{row}$ indicates applying the softmax function along each row of matrix $\mathbf{Q}\mathbf{K}^{\mathrm{T}}$. The $\rho(\mathbf{Q}\mathbf{K}^{\mathrm{T}})$ models the similarities between each pair of pixels of the input, thoroughly extracting the global contextual information contained in the features. However, as $\mathbf{Q} \in \mathbb{R}^{N \times D_k}$ and $\mathbf{K}^{\mathrm{T}} \in \mathbb{R}^{D_k \times N}$, the product between $\mathbf{Q}$ and $\mathbf{K}^{\mathrm{T}}$ belongs to $\mathbb{R}^{N \times N}$, which leads to $O(N^2)$ memory and computation complexity. Therefore, the high resource-demanding of dot-product crucially limits the application on large inputs. Under the condition of softmax normalization function, the $i$-th row of result matrix generated by the dot-product attention module according to equation (9) can be written as:

$$D(\mathbf{Q}, \mathbf{K}, \mathbf{V})_i = \frac{\sum_{j=1}^{N} e^{q_i^T k_j} v_j}{\sum_{j=1}^{N} e^{q_i^T k_j}}, \tag{11}$$

In our previous work on the linear attention (LA) mechanism [3], we design the LA based on first-order approximation of Taylor expansion on equation (11):

$$e^{q_i^T k_j} \approx 1 + \left(\frac{q_i}{\|q_i\|_2}\right)^T \left(\frac{k_j}{\|k_j\|_2}\right). \tag{12}$$

where $l_2$ norm is utilized to ensure $q_i^T k_j \geq -1$. Then, equation (11) can be rewritten as:

$$D(\mathbf{Q}, \mathbf{K}, \mathbf{V})_i = \frac{\sum_{j=1}^{N} \left(1 + \left(\frac{q_i}{\|q_i\|_2}\right)^T \left(\frac{k_j}{\|k_j\|_2}\right)\right) v_j}{\sum_{j=1}^{N} \left(1 + \left(\frac{q_i}{\|q_i\|_2}\right)^T \left(\frac{k_j}{\|k_j\|_2}\right)\right)}, \tag{13}$$

and be simplified as:

$$D(\mathbf{Q}, \mathbf{K}, \mathbf{V})_i = \frac{\sum_{j=1}^{N} v_j + \left(\frac{q_i}{\|q_i\|_2}\right)^T \sum_{j=1}^{N} \left(\frac{k_j}{\|k_j\|_2}\right) v_j^T}{N + \left(\frac{q_i}{\|q_i\|_2}\right)^T \sum_{j=1}^{N} \left(\frac{k_j}{\|k_j\|_2}\right)}. \tag{14}$$

The above equation can be transformed in a vectorized form as:

$$D(\mathbf{Q}, \mathbf{K}, \mathbf{V}) = \frac{\sum_j \mathbf{V}_{i,j} + \left(\frac{\mathbf{Q}}{\|\mathbf{Q}\|_2}\right) \left(\left(\frac{\mathbf{K}}{\|\mathbf{K}\|_2}\right)^T \mathbf{V}\right)}{N + \left(\frac{\mathbf{Q}}{\|\mathbf{Q}\|_2}\right) \sum_j \left(\frac{\mathbf{K}}{\|\mathbf{K}\|_2}\right)^T_{i,j}}. \tag{15}$$

As $\sum_{j=1}^{N} \left(\frac{k_j}{\|k_j\|_2}\right) v_j^T$ and $\sum_{j=1}^{N} \left(\frac{k_j}{\|k_j\|_2}\right)$ can be calculated and reused for every query, time and memory complexity of the proposed LA based on equation (15) is $O(N)$, while the illustration can be seen in Figure 5.

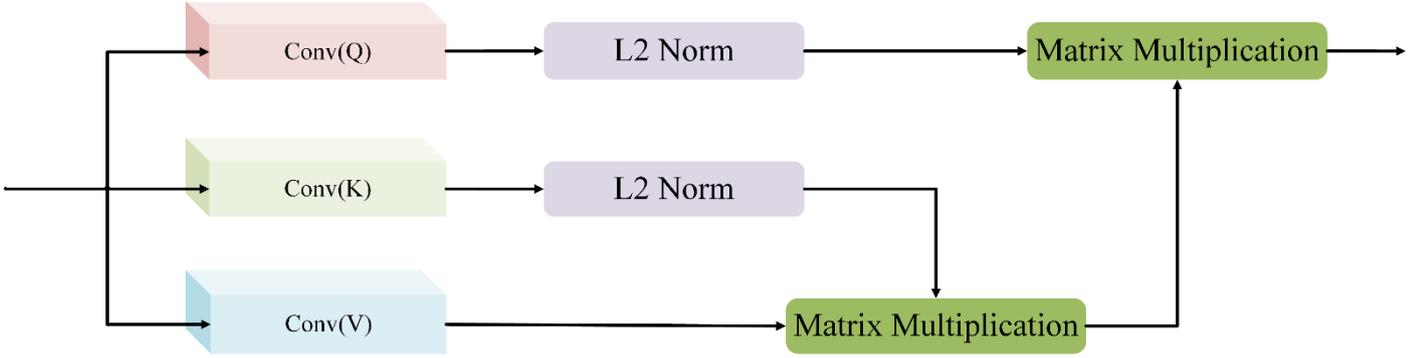

**Figure 5.** The linear attention.

In the FAM, we first employ LA to enhance the spatial relationships of AF, thereby suppressing the fitting residual. Then, a convolutional layer with BN and ReLU is deployed to obtain the attention map. Finally, we apply a matrix multiplication operation between AF and the attention map to obtain the attentional AF. The pipeline of LAM is defined as:

$$\text{LAM}(\mathbf{X}) = \text{Conv}\Big(\text{BN}\Big(\text{ReLU}(\text{LA}(\mathbf{X}))\Big)\Big) \tag{16}$$

Here, Conv represents a standard convolution with a stride of 1.

*Attentional embedding module*: The AEM adopts the LAM to enhance the spatial relationships of LDF4. Then, we apply a matrix multiplication operation between the upsampling attention map of LDF4 and LDF3 to produce the attentional LDF3. Finally, we use an addition operation to fuse the original LDF3 and the attentional LDF3. The pipeline of AEM is illustrated in Figure 6 and can be formalized as:

$$\text{AEM}(\mathbf{LDF3}, \mathbf{LDF4}) = \mathbf{LDF3} + \mathbf{LDF3} \cdot \text{U}\big(\text{LAM}(\mathbf{LDF4})\big) \tag{17}$$

where U denotes the nearest upsampling operation with a scale factor of 2.

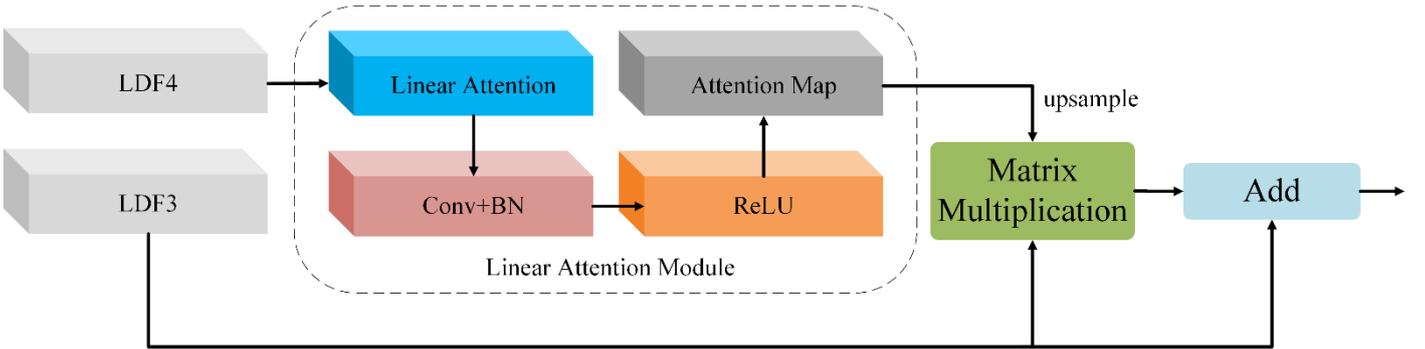

**Figure 6.** The attentional embedding module.

Capitalizing on the benefits provided by feature fusion, the final segmentation feature is abundant in both long-range dependency and textural information for precise semantic segmentation of urban scene images. Besides, linear attention reduces the fitting residual, strengthening the generalization of the network.

## 3. Experiments

In this section, experiments are conducted on three publicly available datasets to evaluate the effectiveness of the proposed BANet. We not only compare the performance of our model on the ISPRS Vaihingen and Potsdam datasets (http://www2.isprs.org/commissions/comm3/wg4/semantic-labeling.html) against the state-of-the-art models designed for remote sensing images but also take those proposed for natural images into consideration. Further, the UAVid dataset [56] is utilized to demonstrate the advantages of our method. Please note that as the backbone for the dependency path of our BANet is ResT-Lite with 10.49 M parameters, the backbone for comparative methods is selected as ResNet-18 with 11.7 M parameters correspondingly for a fair comparison.

### 3.1. Experiments on the ISPRS Vaihingen and Potsdam datasets

#### 3.1.1. Datasets

**Vaihingen**: There are 33 VFR images with a 2494 × 2064 average size in the Vaihingen dataset. The ground sampling distance (GSD) of tiles in Vaihingen is 9 cm. We utilize tiles: 2, 4, 6, 8, 10, 12, 14, 16, 20, 22, 24, 27, 29, 31, 33, 35, 38 for testing, tile: 30 for validation, and the remaining 15 images for training. Please note that we use only the near-infrared, red, and green channels in our experiments. The example images and labels can be seen in the top part of Figure 7.

**Potsdam**: There are 38 fine-resolution images that cover urban scenes in the size of 6000 × 6000 pixels with a 5 cm GSD. We utilize ID: 2_13, 2_14, 3_13, 3_14, 4_13, 4_14, 4_15, 5_13, 5_14, 5_15, 6_13, 6_14, 6_15, 7_13 for testing, ID: 2_10 for validation, and the remaining 22 images, except image named 7_10 with error annotations, for training. Only the red, green, and blue channels are used in our experiments. The example images and labels can be seen in the bottom part of Figure 7.

#### 3.1.2. Training setting

For optimizing the network, the Adam is set as the optimizer with the 0.0003 learning rate and 8 batch size. The images, as well as corresponding labels, are cropped into patches with 512 × 512 pixels and augmented by rotating, resizing, and flipping during training. All the experiments are implemented on a single NVIDIA RTX 3090 GPU with 24 GB RAM. The cross-entropy loss function is utilized as the loss function to measure the disparity between the achieved segmentation maps and the ground reference. If OA on the validation set does not increase for more than 10 epochs, the training procedure will be stopped, while the maximum iteration period is 100 epochs.

#### 3.1.3. Evaluation metrics

The performance of BANet on the ISPRS Potsdam dataset is evaluated using the overall accuracy (OA), the mean Intersection over Union (mIoU), and the F1 score (F1), which are computed on the accumulated confusion matrix:

$$OA = \frac{\sum_{k=1}^{N} TP_k}{\sum_{k=1}^{N} TP_k + FP_k + TN_k + FN_k}, \tag{18}$$

$$mIoU = \frac{1}{N} \sum_{k=1}^{N} \frac{TP_k}{TP_k + FP_k + FN_k}, \tag{19}$$

$$F1 = 2 \times \frac{precision \times recall}{precision + recall}, \tag{20}$$

where $TP_k$, $FP_k$, $TN_k$, and $FN_k$ indicate the true positive, false positive, true negative, and false negatives, respectively, for object indexed as class k. OA is calculated for all categories including the background.

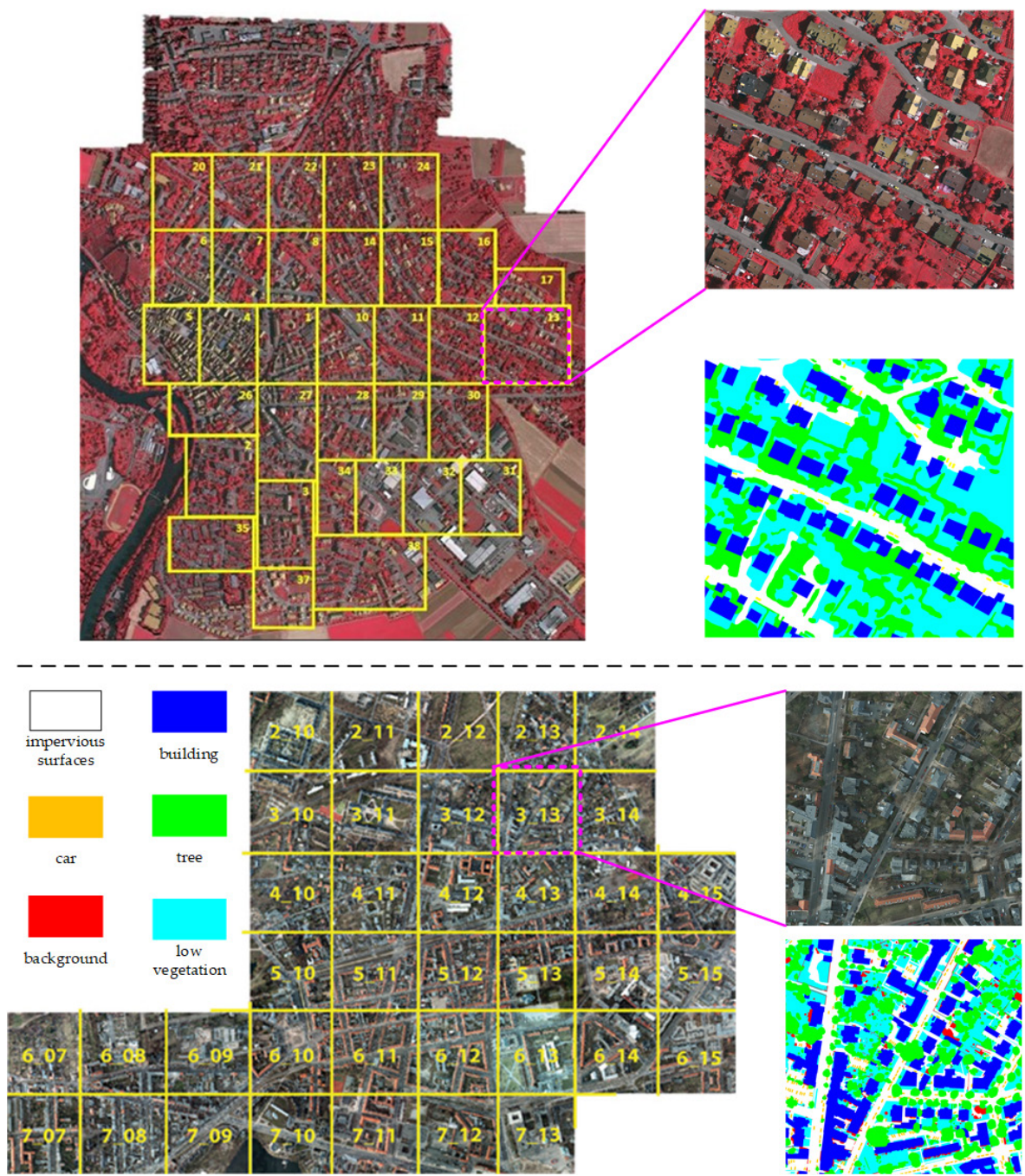

**Figure 7**. Example images and labels from the ISPRS Vaihingen dataset (top part) and Potsdam dataset (bottom part).

### 3.1.4. Experimental results

A detailed comparison between our BANet and other architectures including BiSeNet [57], FANet [58], MAResU-Net [3], EaNet [40], SwiftNet [59], and ShelfNet [60] can be seen in Table 1 and Table 2, based upon the F1-score for each category, mean F1-score, and the OA, and the mIoU on the Vaihingen Potsdam test sets. As it can be observed from the table, the proposed BANet transcends the previous methods designed for segmentation by a large margin, achieving the highest OA of 90.48% and mIoU of 81.35% in the Vaihingen dataset, while the figures for the Potsdam dataset are 91.06% and 86.25%, respectively. Specifically, on the Vaihingen dataset, the proposed BANet brings more than 0.4% improvement in OA and 1.7% improvement in mIoU compared with the suboptimal method, while the improvements for the Potsdam dataset are more than 1.1% and 1.8%.

Particularly, as the relatively small objects, the Car is difficult to recognize in the Vaihingen dataset. Even so, the proposed BANet achieves an 86.76% F1-score, preceding the suboptimal method by more than 5.5%.

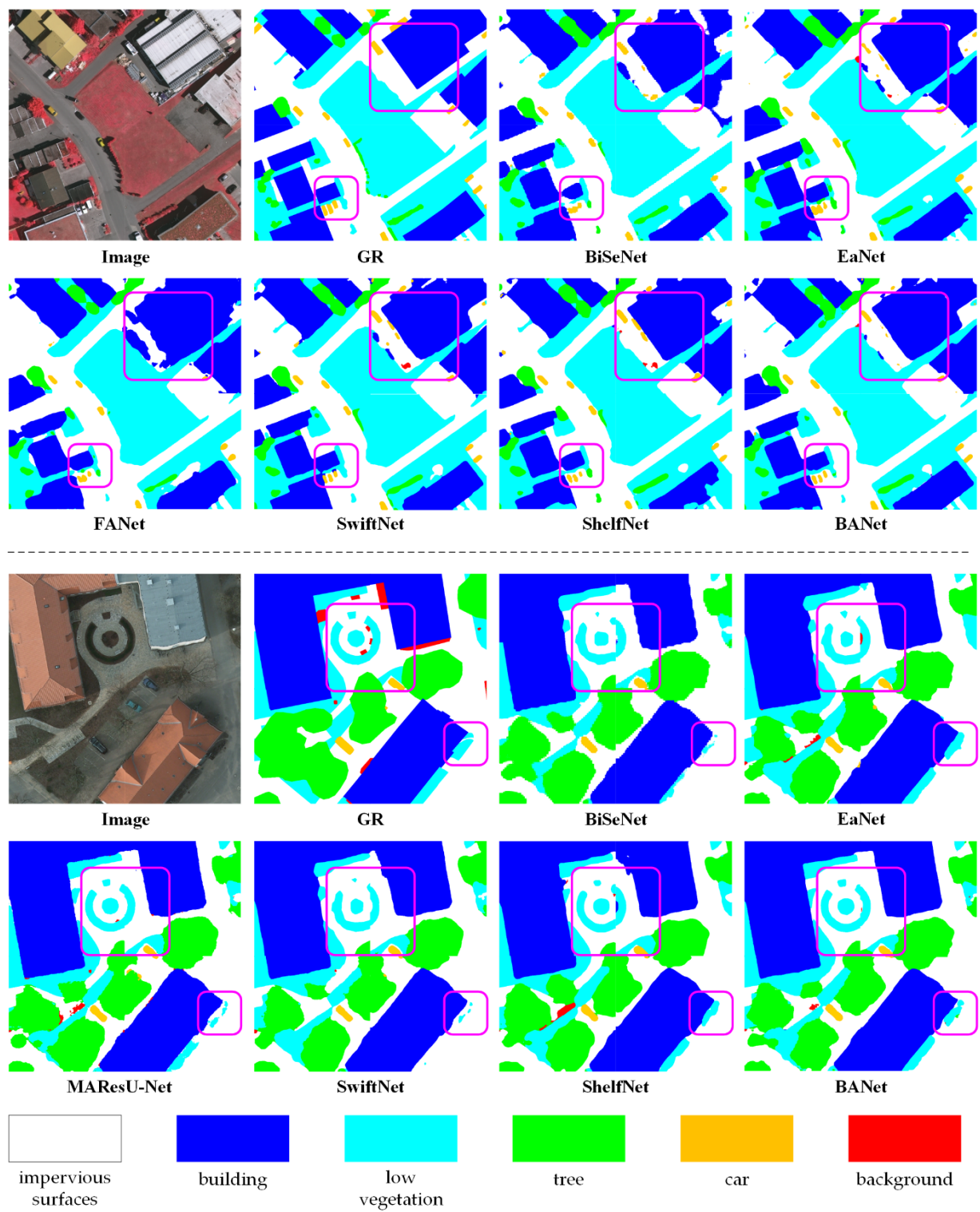

impervious surfaces | building | low vegetation | tree | car | background

**Figure 8**. The experimental results on the ISPRS Vaihingen dataset (top part) and Potsdam dataset (bottom part). GR represents Ground Reference.

To qualitatively validate the effectiveness, we visualize the segmentation maps generated by our BANet and comparative methods in Figure 8. Due to the limited receptive field, the BiSeNet, EaNet, and SwiftNet assign the classification of a specific pixel only by considering a few adjacent areas, leading to fragmented maps and confusion of objects. The direct utilization of the attention mechanism (i.e., MAResU-Net) and the structure of multiple encoder-decoder (i.e., ShelfNet) brings certain improvements. However, the issue of the receptive field is still not entirely resolved. By contrast, we construct the dependency path in our BANet based on an attention-based backbone, i.e., ResT, to capture the long-range global relations, thereby tackling the limitation of the receptive field. Furthermore, a texture path built on convolution operation is equipped in our BANet to utilize the spatial details information in feature maps. Particularly, as shown in Figure 8, the complex circular contour of the Low vegetation is preserved completely by our BANet. Besides, the outlines of the Building generated by our BANet are smoother than those obtained by comparative methods.

**Table 1.** The experimental results on the Vaihingen dataset.

| Method | Backbone | Imp. surf. | Building | Low veg. | Tree | Car | Mean F1 | OA | mIoU |
|--------|----------|-----------|----------|----------|------|-----|---------|-----|------|
| BiSeNet | ResNet-18 | 89.12 | 91.30 | 80.87 | 86.91 | 73.12 | 84.26 | 87.08 | 75.82 |
| FANet | ResNet-18 | 90.65 | 93.78 | 82.60 | 88.56 | 71.60 | 85.44 | 88.87 | 75.61 |
| MAResU-Net | ResNet-18 | 91.97 | 95.04 | 83.74 | 89.35 | 78.28 | 87.68 | 90.07 | 78.58 |
| EaNet | ResNet-18 | 91.68 | 94.52 | 83.10 | 89.24 | 79.98 | 87.70 | 89.69 | 78.68 |
| SwiftNet | ResNet-18 | 92.22 | 94.84 | **84.14** | 89.31 | 81.23 | 88.35 | 90.20 | 79.58 |
| ShelfNet | ResNet-18 | 91.83 | 94.56 | 83.78 | 89.27 | 77.91 | 87.47 | 89.81 | 78.94 |
| BANet | ResT-Lite | **92.23** | **95.23** | 83.75 | **89.92** | **86.76** | **89.58** | **90.48** | **81.35** |

**Table 2.** The experimental results on the Potsdam dataset.

| Method | Backbone | Imp. surf. | Building | Low veg. | Tree | Car | Mean F1 | OA | mIoU |
|--------|----------|-----------|----------|----------|------|-----|---------|-----|------|
| BiSeNet | ResNet-18 | 90.24 | 94.55 | 85.53 | 86.20 | 92.68 | 89.84 | 88.16 | 81.72 |
| FANet | ResNet-18 | 91.99 | 96.10 | 86.05 | 87.83 | 94.53 | 91.30 | 89.82 | 84.16 |
| MAResU-Net | ResNet-18 | 91.41 | 95.57 | 85.82 | 86.61 | 93.31 | 90.54 | 89.04 | 83.87 |
| EaNet | ResNet-18 | 92.01 | 95.69 | 84.31 | 85.72 | 95.11 | 90.57 | 88.70 | 83.38 |
| SwiftNet | ResNet-18 | 91.83 | 95.94 | 85.72 | 86.84 | 94.46 | 90.96 | 89.33 | 83.84 |
| ShelfNet | ResNet-18 | 92.53 | 95.75 | 86.60 | 87.07 | 94.59 | 91.31 | 89.92 | 84.38 |
| BANet | ResT-Lite | **93.34** | **96.66** | **87.37** | **89.12** | **95.99** | **92.50** | **91.06** | **86.25** |

### 3.2. Experiments on the UAVid Dataset

#### 3.2.1. Dataset

As a fine-resolution Unmanned Aerial Vehicle (UAV) semantic segmentation dataset, the UAVid dataset ([https://uavid.nl/](https://uavid.nl/)) is focusing on urban street scenes with a 3840 × 2160 resolution. UAVid is a challenging benchmark since the large resolution of images, large-scale variation, and complexities in the scenes. To be specific, there are 420 images in the dataset where 200 are for training, 70 for validation, and the remaining 150 for testing. The example images and labels can be seen in Figure 9.

We adopt the same hyperparameters and data augmentation as those for experiments on ISPRS datasets, except batch size as 4 and the patch size as 1024 × 1024 during training

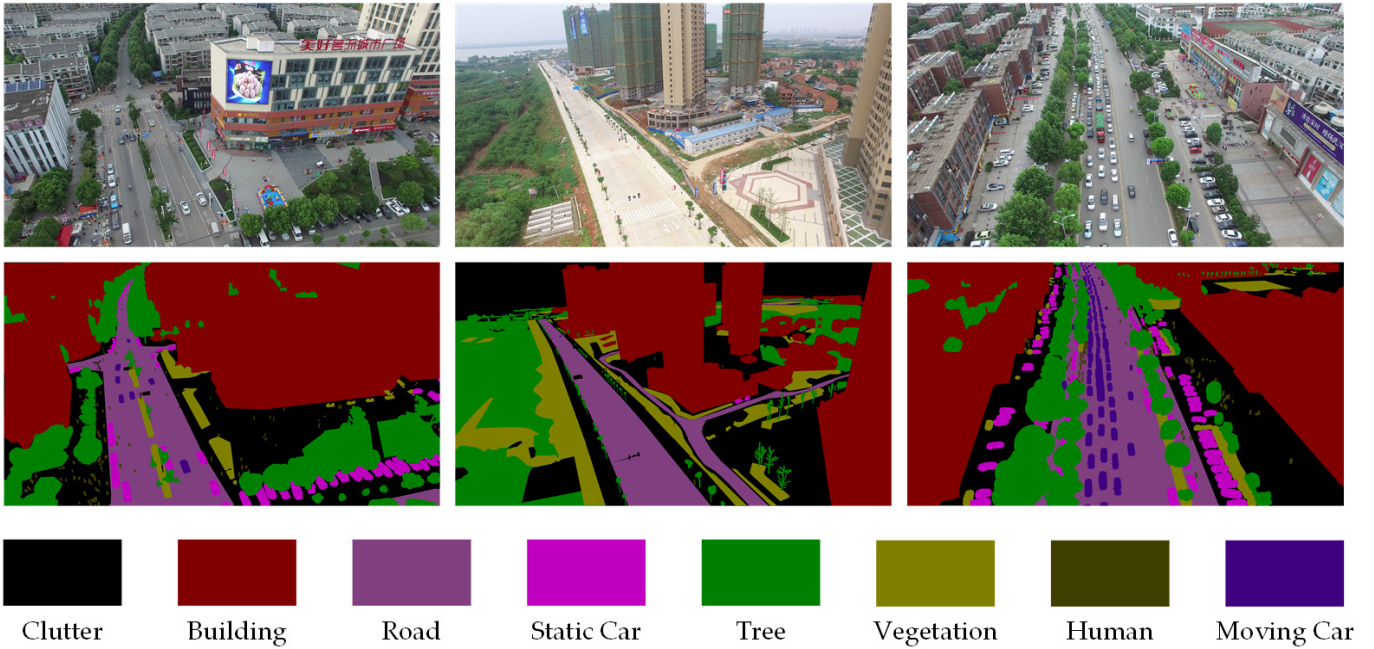

**Figure 9**. Example images and labels from the UAVid dataset.

### 3.2.2. Evaluation metrics

For the UAVid dataset, the performance is assessed from the official server based on the intersection-over-union metric:

$$IoU = \frac{TP_k}{TP_k + FP_k + FN_k},$$

where $TP_k$, $FP_k$, $TN_k$, and $FN_k$ indicate the true positive, false positive, true negative, and false negatives, respectively, for object indexed as class $k$.

### 3.2.3. Experimental results

Quantitative comparison with MSD [56], Fast-SCNN [61], BiSeNet, SwiftNet and ShelfNet are reported in Table 3. As can be seen, the proposed BANet achieves the best IOU score on five out of eight classes, and the best mIoU with a 3% gain over the suboptimal BiSeNet. Qualitative results on the UAVid validation set and test set are demonstrated in Figure 10 and Figure 11, respectively. Compared with the benchmark MSD with obvious local and global inconsistencies, the proposed BANet can effectively capture the cues to scene semantics. For example, in the second row of Figure 11, the cars in the pink box are obviously all moving on the road. However, the MSD identity the left car which is crossing the street as the static car. In contrast, our BANet successfully recognizes all moving cars.

**Table 3**. The experimental results on the UAVid dataset.

| Method | building | tree | clutter | road | vegetation | static car | moving car | human | mIoU |
|--------|----------|------|---------|------|------------|------------|------------|-------|------|
| MSD | 79.8 | 74.5 | 57.0 | 74.0 | 55.9 | 32.1 | 62.9 | 19.7 | 57.0 |
| Fast-SCNN | 75.7 | 71.5 | 44.2 | 61.6 | 43.4 | 19.5 | 51.6 | 0.0 | 45.9 |
| BiSeNet | **85.7** | 78.3 | 64.7 | 61.1 | **77.3** | **63.4** | 48.6 | 17.5 | 61.5 |
| SwiftNet | 85.3 | 78.2 | 64.1 | 61.5 | 76.4 | 62.1 | 51.1 | 15.7 | 61.1 |
| ShelfNet | 76.9 | 73.2 | 44.1 | 61.4 | 43.4 | 21.0 | 52.6 | 3.6 | 47.0 |
| BANet | 85.4 | **78.9** | **66.6** | **80.7** | 62.1 | 52.8 | **69.3** | **21.0** | **64.6** |

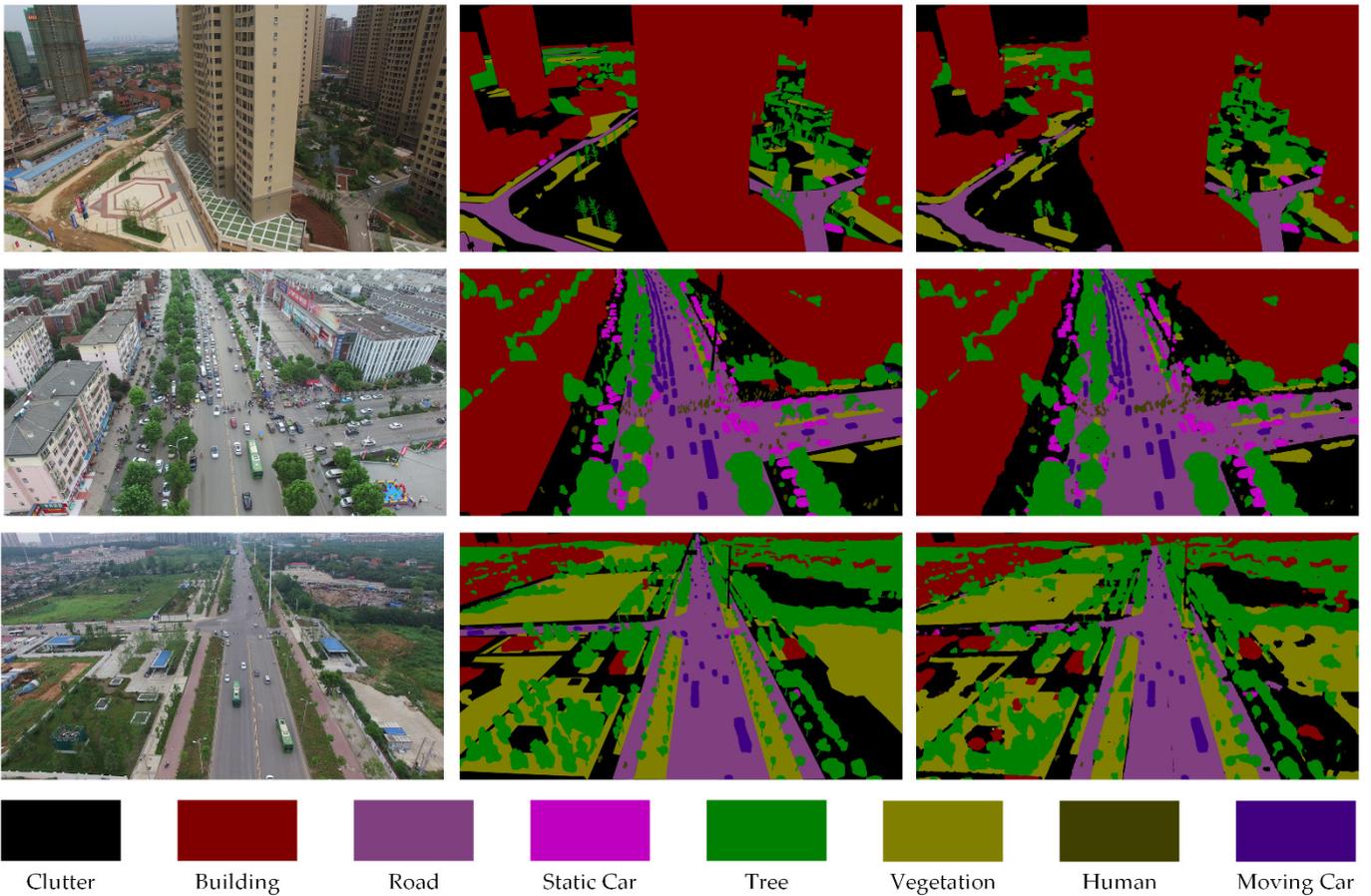

| ■ Clutter | ■ Building | ■ Road | ■ Static Car | ■ Tree | ■ Vegetation | ■ Human | ■ Moving Car |

**Figure 10.** The experimental results on the UAVid validation set. The first column illustrates the input RGB images, the second column depicts the ground reference and the third column shows the predictions of our BANet.

## 4. Discussion

### 4.1. Ablation study

In this part, we conduct extensive ablation experiments on the ISPRS Potsdam dataset to verify the effectiveness of components in the proposed BANet, while the experimental settings and quantitative comparisons are illustrated in Table 4. The results are reported by the average value and corresponding deviation by three-fold experiments. Qualitative comparisons about the ablation study can be seen in Figure 12.

*Baseline*: We select two baselines in ablation experiments, the dependency path which utilizes the ResNet-18 (denoted as ResNet) as the backbone, and the dependency path which adopts the ResT-Lite (denoted as ResT) as the backbone. The feature maps generated by the dependency path are directly up-sampled to restore the shape for final segmentation.

*Ablation for the texture path*: As rich spatial details are important for segmentation, the texture path conducted on the convolution operation is designed in our BANet for preserving the spatial texture. Table 4 illustrates that even the simple fusion schemes such as summation (indicated as Dp+Tp(Sum)) and concatenation (signified as Dp+Tp(Cat)) to merge the texture information can enhance the performance in OA at least 0.2%.

*Ablation for feature aggregation module*: Given the information obtained by the dependency path and the texture path are in different domains, neither summation nor concatenation is the optimal feature fusion scheme. As shown in Table 4, more than 0.5% improvement in OA brings by our BANet compared with Dp+Tp(Sum) and Dp+Tp(Cat) explains the validity of the proposed feature aggregation module.

*Ablation for ResT-Lite*: Since a novel transformer-based backbone, i.e., ResT, is introduced in our BANet, it is valuable to compare the accuracy between the ResNet and ResT. As illustrated in Table 4, the replacement of the backbone in the dependency path brings more than the 1% improvement in OA. Besides, we substitute the backbone in our BANet with ResNet-18 (denoted as BAResNet) to further evaluate the performance. As can be seen in Table 4, a 1.2% gap in OA illuminates the effectiveness of the ResT-Lite. Note that the number of parameters for BAResNet is 14.77 million (59.0 MB for weights file), while the figure for BANet is 15.44 million (56.4 MB for weights file). The inference speed of BAResNet is 73.2 FPS on a single mid-range GPU card, i.e., 1660Ti, for 512 × 512 input images, while the speed of BANet is 33.2 FPS, both satisfy the requirement of real-time (≥30FPS) scenarios. Please notice that the Nvidia GPU has the specialized optimization for CNN, while the optimization for Transformer is not available now. So, the comparison is not completely fair now.

**Table 4**. The experimental results of the ablation study.

| Method | Imp. surf. | Building | Low veg. | Tree | Car | Mean F1 | OA | mIoU |
|--------|-----------|----------|----------|------|-----|---------|-----|------|
| ResNet | 90.91±0.45 | 95.18±0.35 | 84.86±0.92 | 86.44±0.37 | 94.03±0.63 | 90.28±0.28 | 88.48±0.50 | 82.34±0.55 |
| ResT | 92.01±0.58 | 95.73±0.70 | 85.87±0.58 | 87.24±0.80 | 94.13±0.49 | 91.00±0.60 | 89.63±0.49 | 83.80±0.74 |
| Dp+Tp(Sum) | 92.11±0.48 | 95.63±0.43 | 86.5±0.59 | 87.09±0.97 | 94.44±0.30 | 91.15±0.53 | 89.87±0.63 | 84.15±0.72 |
| Dp+Tp(Cat) | 92.30±0.55 | 95.99±0.53 | 86.18±0.64 | 87.57±1.04 | 94.58±0.77 | 91.32±0.68 | 90.35±0.23 | 85.31±0.75 |
| BAResNet | 92.46±0.21 | 95.37±0.43 | 85.92±0.84 | 87.24±0.70 | 94.79±0.28 | 91.16±0.48 | 89.60±0.54 | 84.07±0.64 |
| BANet | 93.27±0.11 | 96.53±0.15 | 87.19±0.23 | 88.63±0.45 | 95.58±0.44 | 92.24±0.23 | 90.86±0.18 | 85.78±0.46 |

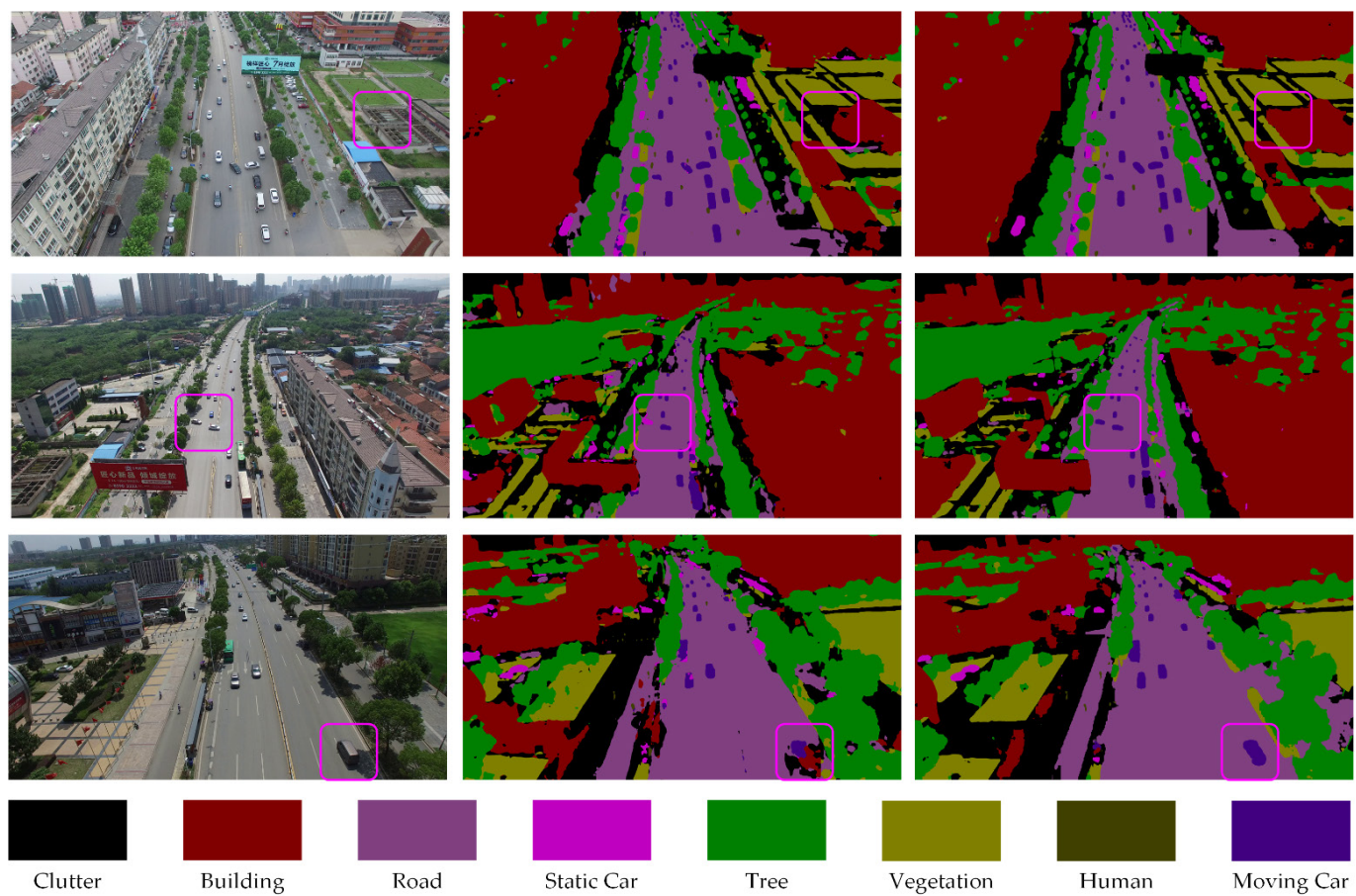

Clutter    Building    Road    Static Car    Tree    Vegetation    Human    Moving Car

**Figure 11**. The experimental results on the UAVid test set. The first column illustrates the input RGB images, the second column depicts the outputs of MSD and the third column shows the predictions of our BANet.

## 4.2. Application scenarios

The main application scenario of our method is urban scene segmentation using remotely sensed images captured by satellite, aerial sensors and UAV drones. The proposed Bilateral Awareness Network, which consists of a texture path, a dependency path and a feature aggregation module, provides a unified framework for semantic segmentation, object detection, and change detection. Moreover, our model considers both accuracy and complexity, revealing enormous potential in illegal land use detection, real-time traffic monitoring, and urban environmental assessment.

In the future, we will continue to study the hybrid structure of convolution and Transformer and apply it to a wider range of urban applications

## 5. Conclusions

This paper proposes a Bilateral Awareness Network for semantic segmentation of very fine resolution urban scene images. Specifically, there are two branches in our BANet, a dependency path built on the Transformer backbone to capture the long-range relationships and a texture path constructed on the convolution operation to exploit the fine-grained details in VHR images. In particular, we further design an attentional feature aggregation module to fuse the global relationship information captured by the dependency path and the spatial texture information generated by the texture path. Extensive experiments on the ISPRS Vaihingen dataset, ISPRS Potsdam dataset, and UAVid dataset demonstrate the effectiveness of the proposed BANet. As a novel exploration to combine the Transformer and convolution in a bilateral structure, we envisage this pioneering paper could inspire practitioners and researchers engaged in this area to explore more possibilities of the Transformer in the remote sensing domain.

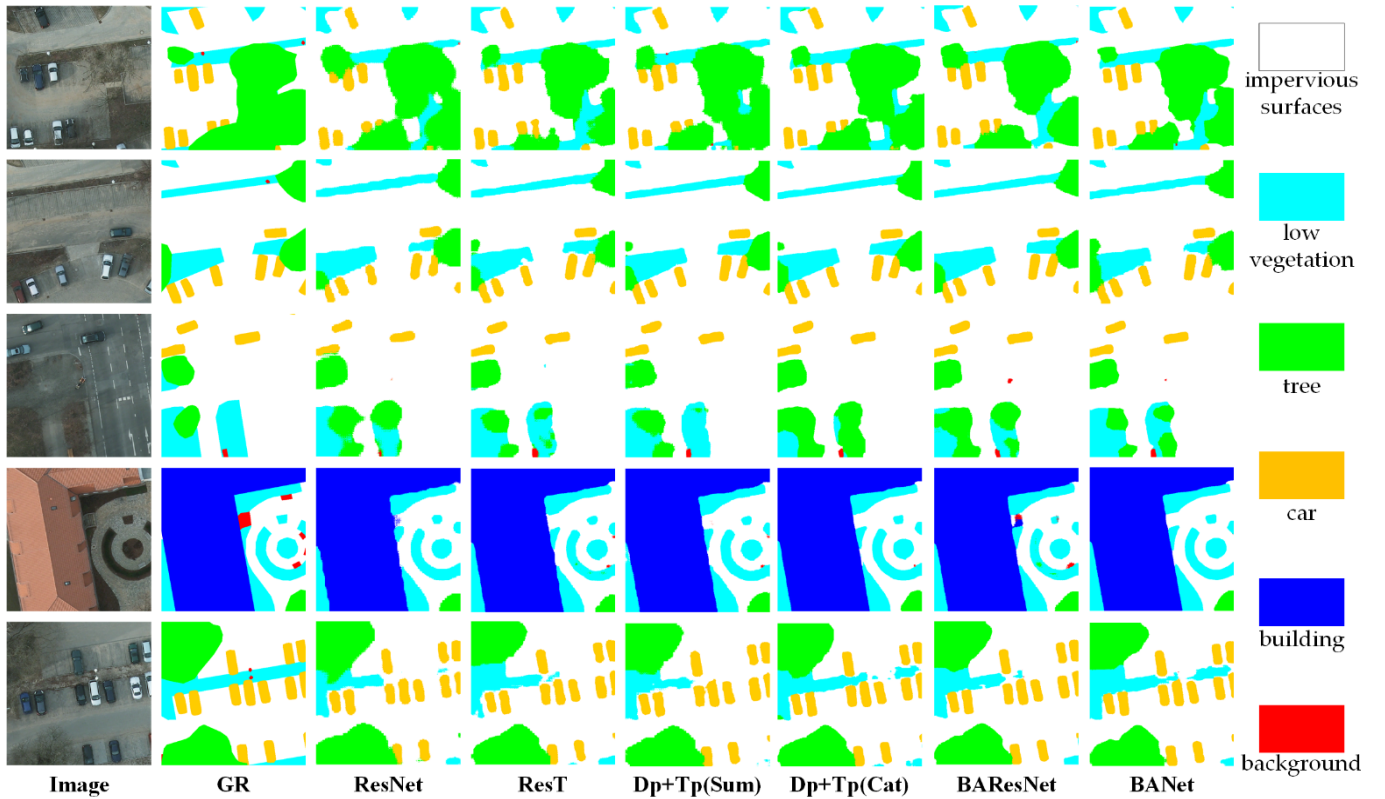

**Figure 12.** The ablation study on the ISPRS Potsdam dataset. GR represents Ground Reference.

**Author Contributions:** This work was conducted in collaboration with all authors. Dongzhi Wang and Teng Wang defined the research theme. Xiaoliang Meng supervised the research work and provided experimental facilities. Libo Wang and Rui Li designed the semantic segmentation model and conducted the experiments. Chenxi Duan checked the experimental results. This manuscript was written by Libo Wang and Rui Li. All authors have read and agreed to the published version of the manuscript.

**Funding:** This research was funded by the National Natural Science Foundation of China (NSFC) under grant number 41971352, National Key Research and Development Program of China under grant number 2018YFB0505003.

**Acknowledgements**: The authors are very grateful to the many people who helped to comment on the article, and the Large Scale Environment Remote Sensing Platform (Facility No. 16000009, 16000011, 16000012) provided by Wuhan University, and the supports provided by Surveying and Mapping Institute Lands and Resource Department of Guangdong Province, Guangzhou. Special thanks to editors and reviewers for providing valuable insight into this article.

**Data Availability Statement:** We are grateful to ISPRS for providing the open benchmarks for 2D remote sensing image semantic segmentation. The data in the paper can be obtained through the following link. Potsdam: https://www2.isprs.org/commissions/comm2/wg4/benchmark/2d-sem-label-potsdam/ Vaihingen: https://www2.isprs.org/commissions/comm2/wg4/benchmark/2d-sem-label-vaihingen/ and UAVid: https://uavid.nl/.

**Conflicts of Interest:** The authors declare no conflict of interest.

**Abbreviations:**

The following abbreviations are used in this manuscript:

| | |
|---|---|
| VFR | Very Fine Resolution |
| DCNNs | Deep Convolutional Neural Networks |
| FCN | Fully Convolutional Neural Network |
| SVM | Support Vector Machine |
| RF | Random Forest |
| CRF | Conditional Random Field |
| MHSA | Multi-Head Self-Attention |
| MLP | Multilayer Perceptron |
| FAM | Feature Aggregation Module |
| BANet | Bilateral Awareness Network |
| TF | Textural Features |
| AF | Aggregated Feature |
| LDF | Long-range Dependent Features |
| BN | Batch Normalization |
| GSD | Ground Sampling Distance |
| UAV | Unmanned Aerial Vehicle |
| LA | Linear Attention |
| AEM | Attentional Embedding Module |
| EMSA | Efficient Multi-head Self-attention |

**References**

1.  Zhang, C.; Atkinson, P.M.; George, C.; Wen, Z.; Diazgranados, M.; Gerard, F. Identifying and mapping individual plants in a highly diverse high-elevation ecosystem using UAV imagery and deep learning. *ISPRS Journal of Photogrammetry and Remote Sensing* **2020**, *169*, 280-291, doi:https://doi.org/10.1016/j.isprsjprs.2020.09.025.

2.  Zhang, C.; Harrison, P.A.; Pan, X.; Li, H.; Sargent, I.; Atkinson, P.M. Scale Sequence Joint Deep Learning (SS-JDL) for land use and land cover classification. *Remote Sensing of Environment* **2020**, *237*, 111593, doi:https://doi.org/10.1016/j.rse.2019.111593.

3.      Li, R.; Zheng, S.; Duan, C.; Su, J.; Zhang, C. Multistage Attention ResU-Net for Semantic Segmentation of Fine-Resolution Remote Sensing Images. *IEEE Geoscience and Remote Sensing Letters* **2021**, 10.1109/LGRS.2021.3063381, 1-5, doi:10.1109/LGRS.2021.3063381.

4.      Li, R.; Duan, C.; Zheng, S.; Zhang, C.; Atkinson, P.M. MACU-Net for Semantic Segmentation of Fine-Resolution Remotely Sensed Images. *IEEE Geoscience and Remote Sensing Letters* **2021**, 10.1109/LGRS.2021.3052886, 1-5, doi:10.1109/LGRS.2021.3052886.

5.      Wang, L.; Fang, S.; Zhang, C.; Li, R.; Duan, C.; Meng, X.; Atkinson, P.M. SaNet: Scale-aware Neural Network for Semantic Labelling of Multiple Spatial Resolution Aerial Images. *arXiv preprint arXiv:2103.07935* **2021**.

6.      Huang, Z.; Wei, Y.; Wang, X.; Shi, H.; Liu, W.; Huang, T.S. AlignSeg: Feature-Aligned Segmentation Networks. *IEEE Transactions on Pattern Analysis and Machine Intelligence* **2021**, 10.1109/TPAMI.2021.3062772, 1-1, doi:10.1109/TPAMI.2021.3062772.

7.      Yao, H.; Qin, R.; Chen, X. Unmanned Aerial Vehicle for Remote Sensing Applications—A Review. *Remote Sensing* **2019**, *11*, doi:10.3390/rs11121443.

8.      Audebert, N.; Le Saux, B.; Lefèvre, S. Segment-before-Detect: Vehicle Detection and Classification through Semantic Segmentation of Aerial Images. *Remote Sensing* **2017**, *9*, doi:10.3390/rs9040368.

9.      Matikainen, L.; Karila, K. Segment-Based Land Cover Mapping of a Suburban Area—Comparison of High-Resolution Remotely Sensed Datasets Using Classification Trees and Test Field Points. *Remote Sensing* **2011**, *3*, doi:10.3390/rs3081777.

10.     Zhang, Q.; Seto, K.C. Mapping urbanization dynamics at regional and global scales using multi-temporal DMSP/OLS nighttime light data. *Remote Sensing of Environment* **2011**, *115*, 2320-2329, doi:https://doi.org/10.1016/j.rse.2011.04.032.

11.     Wei, Y.; Wang, Z.; Xu, M. Road Structure Refined CNN for Road Extraction in Aerial Image. *IEEE Geoscience and Remote Sensing Letters* **2017**, *14*, 709-713, doi:10.1109/LGRS.2017.2672734.

12.     Li, E.; Femiani, J.; Xu, S.; Zhang, X.; Wonka, P. Robust Rooftop Extraction From Visible Band Images Using Higher Order CRF. *IEEE Transactions on Geoscience and Remote Sensing* **2015**, *53*, 4483-4495, doi:10.1109/TGRS.2015.2400462.

13.     Zhang, Y.; Wang, C.; Ji, Y.; Chen, J.; Deng, Y.; Chen, J.; Jie, Y. Combining Segmentation Network and Nonsubsampled Contourlet Transform for Automatic Marine Raft Aquaculture Area Extraction from Sentinel-1 Images. *Remote Sensing* **2020**, *12*, 4182.

14.     Maxwell, A.E.; Bester, M.S.; Guillen, L.A.; Ramezan, C.A.; Carpinello, D.J.; Fan, Y.; Hartley, F.M.; Maynard, S.M.; Pyron, J.L. Semantic Segmentation Deep Learning for Extracting Surface Mine Extents from Historic Topographic Maps. *Remote Sensing* **2020**, *12*, 4145.

15.     Kalajdjieski, J.; Zdravevski, E.; Corizzo, R.; Lameski, P.; Kalajdziski, S.; Pires, I.M.; Garcia, N.M.; Trajkovik, V. Air pollution prediction with multi-modal data and deep neural networks. *Remote Sensing* **2020**, *12*, 4142.

16.     Diakogiannis, F.I.; Waldner, F.; Caccetta, P.; Wu, C. ResUNet-a: A deep learning framework for semantic segmentation of remotely sensed data. *ISPRS Journal of Photogrammetry and Remote Sensing* **2020**, *162*, 94-114, doi:https://doi.org/10.1016/j.isprsjprs.2020.01.013.

17.     Li, R.; Duan, C. ABCNet: Attentive Bilateral Contextual Network for Efficient Semantic Segmentation of Fine-Resolution Remote Sensing Images. *arXiv preprint arXiv:2102.02531* **2021**.

18.     Zhang, C.; Sargent, I.; Pan, X.; Li, H.; Gardiner, A.; Hare, J.; Atkinson, P.M. Joint Deep Learning for land cover and land use classification. *Remote Sensing of Environment* **2019**, *221*, 173-187, doi:https://doi.org/10.1016/j.rse.2018.11.014.

19.     Zhang, C.; Sargent, I.; Pan, X.; Li, H.; Gardiner, A.; Hare, J.; Atkinson, P.M. An object-based convolutional neural network (OCNN) for urban land use classification. *Remote Sensing of Environment* **2018**, *216*, 57-70, doi:https://doi.org/10.1016/j.rse.2018.06.034.

20.    Long, J.; Shelhamer, E.; Darrell, T. Fully convolutional networks for semantic segmentation. In Proceedings of IEEE/CVF Conference on Computer Vision and Pattern Recognition, 2015; pp. 3431-3440.

21.    Sherrah, J. Fully convolutional networks for dense semantic labelling of high-resolution aerial imagery. *arXiv preprint arXiv:1606.02585* **2016**.

22.    Guo, Y.; Jia, X.; Paull, D. Effective Sequential Classifier Training for SVM-Based Multitemporal Remote Sensing Image Classification. *IEEE Transactions on Image Processing* **2018**, *27*, 3036-3048, doi:10.1109/TIP.2018.2808767.

23.    Pal, M. Random forest classifier for remote sensing classification. *International Journal of Remote Sensing* **2005**, *26*, 217-222, doi:10.1080/01431160412331269698.

24.    Krähenbühl, P.; Koltun, V. Efficient inference in fully connected crfs with gaussian edge potentials. *Advances in neural information processing systems* **2011**, *24*, 109-117.

25.    Ma, L.; Liu, Y.; Zhang, X.; Ye, Y.; Yin, G.; Johnson, B.A. Deep learning in remote sensing applications: A meta-analysis and review. *ISPRS Journal of Photogrammetry and Remote Sensing* **2019**, *152*, 166-177, doi:https://doi.org/10.1016/j.isprsjprs.2019.04.015.

26.    Marcos, D.; Volpi, M.; Kellenberger, B.; Tuia, D. Land cover mapping at very high resolution with rotation equivariant CNNs: Towards small yet accurate models. *ISPRS Journal of Photogrammetry and Remote Sensing* **2018**, *145*, 96-107, doi:https://doi.org/10.1016/j.isprsjprs.2018.01.021.

27.    Yue, K.; Yang, L.; Li, R.; Hu, W.; Zhang, F.; Li, W. TreeUNet: Adaptive Tree convolutional neural networks for subdecimeter aerial image segmentation. *ISPRS Journal of Photogrammetry and Remote Sensing* **2019**, *156*, 1-13, doi:https://doi.org/10.1016/j.isprsjprs.2019.07.007.

28.    Ronneberger, O.; Fischer, P.; Brox, T. U-Net: Convolutional Networks for Biomedical Image Segmentation. In Proceedings of Medical Image Computing and Computer-Assisted Intervention – MICCAI 2015, Cham, 2015//; pp. 234-241.

29.    Liu, Y.; Fan, B.; Wang, L.; Bai, J.; Xiang, S.; Pan, C. Semantic labeling in very high resolution images via a self-cascaded convolutional neural network. *ISPRS Journal of Photogrammetry and Remote Sensing* **2018**, *145*, 78-95, doi:10.1016/j.isprsjprs.2017.12.007.

30.    Yang, M.Y.; Kumaar, S.; Lyu, Y.; Nex, F. Real-time Semantic Segmentation with Context Aggregation Network. *ISPRS Journal of Photogrammetry and Remote Sensing* **2021**, *178*, 124-134, doi:https://doi.org/10.1016/j.isprsjprs.2021.06.006.

31.    Fu, J.; Liu, J.; Tian, H.; Li, Y.; Bao, Y.; Fang, Z.; Lu, H. Dual attention network for scene segmentation. In Proceedings of IEEE/CVF Conference on Computer Vision and Pattern Recognition; pp. 3146-3154.

32.    Li, R.; Zheng, S.; Zhang, C.; Duan, C.; Su, J.; Wang, L.; Atkinson, P.M. Multiattention Network for Semantic Segmentation of Fine-Resolution Remote Sensing Images. *IEEE Transactions on Geoscience and Remote Sensing* **2021**.

33.    Zhao, H.; Qi, X.; Shen, X.; Shi, J.; Jia, J. Icnet for real-time semantic segmentation on high-resolution images. In Proceedings of European conference on computer vision (ECCV), 2018; pp. 405-420.

34.    Kampffmeyer, M.; Salberg, A.-B.; Jenssen, R. Semantic segmentation of small objects and modeling of uncertainty in urban remote sensing images using deep convolutional neural networks. In Proceedings of IEEE conference on computer vision and pattern recognition workshops; pp. 1-9.

35.    Maggiori, E.; Tarabalka, Y.; Charpiat, G.; Alliez, P. High-Resolution Aerial Image Labeling With Convolutional Neural Networks. *IEEE Transactions on Geoscience and Remote Sensing* **2017**, *55*, 7092-7103, doi:10.1109/TGRS.2017.2740362.

36.    Audebert, N.; Le Saux, B.; Lefèvre, S. Beyond RGB: Very high resolution urban remote sensing with multimodal deep networks. *ISPRS Journal of Photogrammetry and Remote Sensing* **2018**, *140*, 20-32, doi:https://doi.org/10.1016/j.isprsjprs.2017.11.011.

37.    Duan, C.; Pan, J.; Li, R. Thick Cloud Removal of Remote Sensing Images Using Temporal Smoothness and Sparsity Regularized Tensor Optimization. *Remote Sensing* **2020**, *12*, 3446.

38. Kampffmeyer, M.; Salberg, A.B.; Jenssen, R. Urban Land Cover Classification With Missing Data Modalities Using Deep Convolutional Neural Networks. *IEEE Journal of Selected Topics in Applied Earth Observations and Remote Sensing* **2018**, *11*, 1758-1768, doi:10.1109/JSTARS.2018.2834961.

39. Marmanis, D.; Schindler, K.; Wegner, J.D.; Galliani, S.; Datcu, M.; Stilla, U. Classification with an edge: Improving semantic image segmentation with boundary detection. *ISPRS Journal of Photogrammetry and Remote Sensing* **2018**, *135*, 158-172, doi:https://doi.org/10.1016/j.isprsjprs.2017.11.009.

40. Zheng, X.; Huan, L.; Xia, G.-S.; Gong, J. Parsing very high resolution urban scene images by learning deep ConvNets with edge-aware loss. *ISPRS Journal of Photogrammetry and Remote Sensing* **2020**, *170*, 15-28, doi:10.1016/j.isprsjprs.2020.09.019.

41. Wang, X.; Girshick, R.; Gupta, A.; He, K. Non-local neural networks. In Proceedings of Proceedings of the IEEE conference on computer vision and pattern recognition, 2018; pp. 7794-7803.

42. Yu, F.; Koltun, V. Multi-scale context aggregation by dilated convolutions. *arXiv preprint arXiv:1511.07122* **2015**.

43. Liu, Q.; Kampffmeyer, M.; Jenssen, R.; Salberg, A.B. Dense Dilated Convolutions' Merging Network for Land Cover Classification. *IEEE Transactions on Geoscience and Remote Sensing* **2020**, *58*, 6309-6320, doi:10.1109/TGRS.2020.2976658.

44. Huang, Z.; Wang, X.; Wei, Y.; Huang, L.; Shi, H.; Liu, W.; Huang, T.S. CCNet: Criss-Cross Attention for Semantic Segmentation. *IEEE Transactions on Pattern Analysis and Machine Intelligence* **2020**, 10.1109/TPAMI.2020.3007032, 1-1, doi:10.1109/TPAMI.2020.3007032.

45. Vaswani, A.; Shazeer, N.; Parmar, N.; Uszkoreit, J.; Jones, L.; Gomez, A.N.; Kaiser, L.; Polosukhin, I. Attention is all you need. *arXiv preprint arXiv:1706.03762* **2017**.

46. Dosovitskiy, A.; Beyer, L.; Kolesnikov, A.; Weissenborn, D.; Zhai, X.; Unterthiner, T.; Dehghani, M.; Minderer, M.; Heigold, G.; Gelly, S. An image is worth 16x16 words: Transformers for image recognition at scale. *arXiv preprint arXiv:2010.11929* **2020**.

47. Zhu, X.; Su, W.; Lu, L.; Li, B.; Wang, X.; Dai, J. Deformable DETR: Deformable Transformers for End-to-End Object Detection. *arXiv preprint arXiv:2010.04159* **2020**.

48. Wang, L.; Li, R.; Duan, C.; Fang, S. Transformer Meets DCFAM: A Novel Semantic Segmentation Scheme for Fine-Resolution Remote Sensing Images. *arXiv preprint arXiv:2104.12137* **2021**.

49. Ba, J.L.; Kiros, J.R.; Hinton, G.E. Layer normalization. *arXiv preprint arXiv:1607.06450* **2016**.

50. Liu, Z.; Lin, Y.; Cao, Y.; Hu, H.; Wei, Y.; Zhang, Z.; Lin, S.; Guo, B. Swin transformer: Hierarchical vision transformer using shifted windows. *arXiv preprint arXiv:2103.14030* **2021**.

51. Ioffe, S.; Szegedy, C. Batch normalization: Accelerating deep network training by reducing internal covariate shift. In Proceedings of International conference on machine learning; pp. 448-456.

52. Nair, V.; Hinton, G.E. Rectified linear units improve restricted boltzmann machines. In Proceedings of ICML.

53. Zhang, Q.; Yang, Y. ResT: An Efficient Transformer for Visual Recognition. *arXiv preprint arXiv:2105.13677* **2021**.

54. Chollet, F. Xception: Deep learning with depthwise separable convolutions. In Proceedings of IEEE conference on computer vision and pattern recognition, 2017; pp. 1251-1258.

55. Ulyanov, D.; Vedaldi, A.; Lempitsky, V. Instance normalization: The missing ingredient for fast stylization. *arXiv preprint arXiv:1607.08022* **2016**.

56. Lyu, Y.; Vosselman, G.; Xia, G.-S.; Yilmaz, A.; Yang, M.Y. UAVid: A semantic segmentation dataset for UAV imagery. *ISPRS Journal of Photogrammetry and Remote Sensing* **2020**, *165*, 108-119, doi:https://doi.org/10.1016/j.isprsjprs.2020.05.009.

57. Yu, C.; Wang, J.; Peng, C.; Gao, C.; Yu, G.; Sang, N. Bisenet: Bilateral segmentation network for real-time semantic segmentation. In Proceedings of European conference on computer vision (ECCV) pp. 325-341.

58. Hu, P.; Perazzi, F.; Heilbron, F.C.; Wang, O.; Lin, Z.; Saenko, K.; Sclaroff, S. Real-Time Semantic Segmentation With Fast Attention. *IEEE Robotics and Automation Letters* **2021**, *6*, 263-270, doi:10.1109/LRA.2020.3039744.

59. Oršić, M.; Šegvić, S. Efficient semantic segmentation with pyramidal fusion. *Pattern Recognition* **2021**, *110*, 107611, doi:https://doi.org/10.1016/j.patcog.2020.107611.

60. Zhuang, J.; Yang, J.; Gu, L.; Dvornek, N. Shelfnet for fast semantic segmentation. In Proceedings of IEEE/CVF International Conference on Computer Vision Workshops; pp. 0-0.

61. Poudel, R.P.K.; Liwicki, S.; Cipolla, R. Fast-scnn: Fast semantic segmentation network. *arXiv preprint arXiv:1902.04502* **2019**.